\documentclass[11pt,a4paper]{article}
\usepackage[hyperref]{acl2019}
\usepackage{times}
\usepackage{latexsym}
\usepackage{todonotes}

\usepackage{graphicx}
\usepackage{url}
\usepackage{enumitem}

\aclfinalcopy 

\title{Advancing NLP with Cognitive Language Processing Signals}

\author{Nora Hollenstein \\
  ETH Zurich \\
  \texttt{noraho@ethz.ch} \\\And
  Maria Barrett \\
  University of Copenhagen \\
  \texttt{mjb@di.ku.dk} \\\And
  Marius Troendle \\
  University of Zurich \\
  \texttt{m.troendle@uzh.ch}\\\AND
  Francesco Bigiolli \\
  ETH Zurich \\
  \texttt{fbigiol@ethz.ch} \\ \And
  Nicolas Langer \\
  University of Zurich \\
  \texttt{n.langer@uzh.ch} \\ \And
  Ce Zhang \\
  ETH Zurich \\
  \texttt{ce.zhang@ethz.ch} \\
  }

\date{}

\begin{document}
\maketitle
\begin{abstract}
  When we read, our brain processes language and generates cognitive processing data such as gaze patterns and brain activity. These signals can be recorded while reading. Cognitive language processing data such as eye-tracking features have shown improvements on single NLP tasks. We analyze whether using such human features can show consistent improvement across tasks and data sources. We present an extensive investigation of the benefits and limitations of using cognitive processing data for NLP. Specifically, we use gaze and EEG features to augment models of named entity recognition, relation classification, and sentiment analysis. These methods significantly outperform the baselines and show the potential and current limitations of employing human language processing data for NLP.
  
\end{abstract}


\section{Introduction}

When reading, humans process language ``automatically'' without reflecting on each step --- Humans string words together into sentences, understand the meaning of spoken and written ideas, and process language without thinking too much about how the underlying cognitive process happens. This process generates cognitive signals that could potentially facilitate natural language processing tasks.

In recent years, collecting these signals
has become increasingly easy and less expensive \cite{papoutsaki2016webgazer};
as a result, using cognitive features to improve NLP tasks has become more popular.
For example, researchers have proposed a range of work that uses eye-tracking or gaze 
signals to improve part-of-speech tagging \citep{barrett2016weakly}, sentiment analysis \citep{mishra2017leveraging}, named entity recognition \cite{hollenstein2019entity}, among other tasks. Moreover, these signals have been used successfully to regularize attention in neural networks for NLP \cite{barrett2018sequence}.

However, most previous work leverages only eye-tracking data, presumably because it is the most accessible form of cognitive language processing signal. In addition, most state-of-the-art work focused on improving a single task with a single type of cognitive signal. But {\em can cognitive processing signal bring consistent improvements across modality (e.g., eye-tracking and/or EEG) and across various NLP tasks?} And if so, {\em does the combination of different sources of cognitive signals bring incremental improvements?}

In this paper, we aim at shedding light on these questions. We present, to the best of our knowledge, the first comprehensive study to analyze the benefits and limitations of using cognitive language processing signals to improve NLP across multiple tasks and modalities (types of signals).
Specifically, we go beyond state-of-the-art in two ways:

\vspace{0.1em}
\noindent
{\bf (Multiple Signals)} We consider
both eye-tracking and electroencephalography (EEG) data as examples of cognitive language processing data. Eye-tracking records the readers gaze positions on the screen and serves as an indirect measure of the cognitive reading process. EEG records electrical brain activity along the scalp and is a more direct measure of physiological processes, including language processing. This is also the first application leveraging EEG data to improve NLP tasks.

\vspace{0.1em}
\noindent
{\bf (Multiple Tasks)}
We then construct named entity recognition, relation classification, and sentiment analysis models with gaze and EEG features. We analyze three methods of adding these cognitive signals to machine learning architectures for NLP. First, we simply add the features to existing systems (Section \ref{sec:tasks}). 
Second, we show how these features can be generalized so that recorded data is not required at test data (Section \ref{sec:type}). And third, in a multi-task setting we learn gaze and EEG features as auxiliary tasks to aid the main NLP task (Section \ref{sec:mtl}).

\vspace{0.1em}
\noindent 
In summary, the most important insights gained from this work include:

\vspace{0.4em}
\noindent
{\bf 1.} Using cognitive features shows consistent improvements over a range of NLP tasks even without large amounts of recorded cognitive signals.
    
\vspace{0.4em}
\noindent
{\bf 2.} While integrating gaze or EEG signals separately significantly outperforms the baselines, the combination of both does not further improve the results.

\vspace{0.4em}
\noindent
{\bf 3.}  We identify multiple directions of future research: {\em How  can  cognitive  signals, such as EEG data,  be  preprocessed and de-noised more efficiently for NLP tasks?  How can cognitive features of different sources be combined more effectively for natural language processing?}

\vspace{0.2em}
\noindent All experiments presented in this paper are available\footnote{\url{https://github.com/DS3Lab/zuco-nlp/}} to provide a foundation for future work to better understand these questions.


\section{Related Work}
\subsection{Eye-tracking}

The benefits of eye movement data have been assessed in various domains, including NLP and computer vision. Eye-trackers provide millisecond-accurate records about where humans look when they are reading. Although it is mostly still being recorded in controlled environments, recent approaches have shown substantial improvements in recording gaze data by using cameras of mobile devices \citep{gomez2016evaluation,papoutsaki2016webgazer}. Hence, gaze data will become more accessible and available in much larger volumes in the next few years \cite{san2009low,sewell2010real}, which will facilitate the creation of sizable datasets enormously.

The benefit of eye-tracking in human language processing 
is supported by intensive study in psycholinguistics during the 20th century and onwards. For example, when humans read a text, they do not focus on every single word. The number of fixations and the fixation duration on a word depends on a number of linguistic factors \cite{clifton2007eye,demberg2008data}. Different features even allow us to study early and late cognitive processing separately. 

First, word length, frequency and predictability from context affect fixation duration and counts. The frequency effect was first noted by \citet{rayner1977visual} and has been consistently reported in various studies since, e.g. \citet{just1980theory,rayner1986lexical,cop2017presenting}. Second, readers are more likely to fixate on open-class words \citep{carpenter1983your}. It even appears that eye movements are reliable indicators of syntactical categories \citep{barrett2015reading}. 

Word familiarity also influences how long readers look at a word. Although two words may have the same frequency value, they may differ in familiarity and predictability from context. Effects of word familiarity on fixation time have also been demonstrated in a number of studies \cite{juhasz2003investigating,williams2004eye} as have word predictability effects, e.g. \citet{mcdonald2003eye}.

A range of work of using eye-tracking signals to improve natural language processing tasks has been proposed and shows promising results. Gaze data has been used to improve tasks such as part-of-speech tagging \citep{barrett2016weakly}, sentiment analysis \citep{mishra2017leveraging}, prediction of multiword expressions \citep{rohanian2017using}, sentence compression \citep{klerke2016improving}, and word embedding evaluation \citep{sogaard2016evaluating}. Furthermore, gaze data has been used to regularize attention in neural architectures on NLP classification tasks \cite{barrett2018sequence}.

\subsection{EEG}
\vspace{-0.2em}
To the best of our knowledge, there are no applications leveraging EEG data to improve NLP tasks. There are, however, good reasons to try to combine the two sources. EEG could provide the missing information in the eye movements to disambiguate different cognitive processes. An extended fixation duration only tells us that extended cognitive processing occurs, but not \emph{which process}. 

EEG and eye-tracking use the same temporal resolution with non-invasive technologies \citep{sereno2003measuring}. \citet{dambacher2007synchronizing} found that longer fixation duration correlates with larger N400 amplitude effects. N400 is part of the normal brain response to words and other meaningful stimuli \cite{kutas2000electrophysiology}. Effects of word predictability on eye movements and EEG co-registration have also been studied in serialized word representation and in natural reading \cite{dimigen2011coregistration}.

Other aspects relevant for linguistic processing can be observed in the EEG signal itself. For instance, term relevance can be associated with brain activity with significant changes in certain brain areas \cite{eugster2014predicting}, differences in processing verbs and noun, concrete nouns and abstract nouns, as well as common nouns and proper nouns are also observed \cite{weiss2003contribution}. Furthermore, there is a correspondence between computational grammar models and certain EEG effects \cite{hale2018finding}.


Collecting EEG data is more expensive and time-consuming than collecting eye-tracking data, which is why brain activity data is commonly less accessible. Moreover, collecting EEG data from subjects in a naturalistic reading environment is even more challenging. Hence, related work in this area is very limited. Subsequently, while we rely on standard practices when leveraging gaze data, our experiments using EEG data are more experimental.


\section{Data}\label{data}

The Zurich Cognitive Language Processing Corpus (ZuCo; \citet{hollenstein2018zuco}) is the main data source of this work. It is the first freely available dataset\footnote{The data is available here: \url{https://osf.io/q3zws/}} of simultaneous eye-tracking and EEG recordings of natural sentence reading. This corpus includes recordings of 12 adult, native speakers reading approximately 1100 English sentences.  

The corpus contains both natural reading and task-solving reading paradigms. For this work, we make use of the first two reading paradigms of ZuCo, during which the subjects read naturally at their own speed and without any specific task other than answering some control questions testing their reading comprehension. The first paradigm includes 300 sentences (7737 tokes) from Wikipedia articles \cite{culotta2006integrating} that contained semantic relations such as \textit{employer}, \textit{award} and \textit{job\_title}. The second paradigm contains 400 positive, negative and neutral sentences (8138 tokens) from the Stanford Sentiment Treebank \cite{socher2013recursive}, to analyze the elicitation of emotions and opinions during reading. The same sentences were read by all 12 subjects.
\vspace{-0.2em}
\subsection{Gaze features}\label{et}

ZuCo readily provides 5 eye-tracking features: number of fixations (NFIX), the number of all fixations landing on a word; first fixation duration (FFD), the duration of the first fixation on the current word; total reading time (TRT), the sum of all fixation durations on the current word; gaze duration (GD), the sum of all fixations on the current word in the first-pass reading before the eye moves out of the word; and go-past time (GPT), the sum of all fixations prior to progressing to the right of the current word, including regressions to previous words that originated from the current word.
Fixations shorter than 100 ms were excluded, since these are unlikely to reflect language processing \cite{sereno2003measuring}. To increase the robustness of the signal, the eye-tracking features are averaged over all subjects.
\vspace{-0.2em}
\subsection{EEG features}

Since eye-tracking and EEG were recorded simultaneously, we were able to extract word-level EEG features. During the preprocessing of ZuCo 23 electrodes in the outermost circumference (chin and neck) were used to detect muscular artifacts and were removed for subsequent analyses. Thus, each EEG feature, corresponding to the duration of a specific fixation, contains 105 electrode values. The EEG signal is split into 8 frequency bands, which are fixed ranges of wave frequencies and amplitudes over a time scale: \textit{theta1} (4-6 Hz), \textit{theta2} (6.5-8 Hz), \textit{alpha1} (8.5-10 Hz), \textit{alpha2} (10.5-13 Hz), \textit{beta1}, (13.5-18 Hz) beta2 (18.5-30 Hz), \textit{gamma1} (30.5-40 Hz) and \textit{gamma2} (40-49.5 Hz). These frequency ranges are known to correlate with certain cognitive functions.  For instance, theta activity reflects cognitive control and working memory \cite{williams2019thinking}, alpha activity has been related to attentiveness \cite{klimesch2012alpha}, gamma-band activity has been used to detect emotions \cite{li2009emotion} and beta frequencies affect decisions regarding relevance \cite{eugster2014predicting}. Even though the variability between subjects is much higher in the EEG signal, we also average all features over all subjects.


\section{Tasks}\label{sec:tasks}
\vspace{-0.2em}
To thoroughly evaluate the potential of gaze and brain activity data, we perform experiments on the three information extraction tasks described in this section. Current state-of-the-art systems are used for all tasks and different combinations of cognitive features are evaluated.

\subsection{Named Entity Recognition}

The performance of named entity recognition (NER) systems can successfully be improved with eye-tracking features \cite{hollenstein2019entity}. However, this has not been explored for EEG signals. We use the state-of-the-art neural architecture for NER by \citet{lample2016neural}\footnote{\url{https://github.com/glample/tagger}}. Their model successfully combines word-level and character-level embeddings, which we augment with embedding layers for gaze and/or EEG features. Word length and frequency are known to correlate and interact with gaze features (e.g. \citet{just1980theory,rayner1977visual}), which is why we selected a base model that allows us to combine the cognitive features with word-level and character-level information. We use the named entity annotations from \url{https://github.com/DS3Lab/ner-at-first-sight}.
\vspace{-0.2em}
\paragraph{Features} For this task, we used the 17 gaze features proposed by \citet{hollenstein2019entity} for NER. These features include relevant information from early and late word processing as well as context features from the surrounding words. We extracted 8 word-level EEG features, one for each frequency band (The neural architecture of this system does not allow for raw normalized EEG and gaze features as is the case for relation classification and sentiment analysis.). The feature values were averaged over the 105 electrode values. These features are mapped to the duration of the gaze features. Thus, in the experiments we tested EEG features during total reading time of the words and EEG features merely during the first fixations. The latter yielded better results. The gaze and EEG features values (originally in milliseconds (for gaze) and microvolts (for EEG)) were normalized and concatenated to the character and word embeddings as one-hot vectors.
\vspace{-0.2em}
\paragraph{Experiments} All models were trained on both ZuCo paradigms described above (15875 tokens) with 10-fold cross validation (80\% training, 10\% development, 10\% test) and early stopping was performed after 20 epochs of no improvement on the development set to reduce training time. For the experiments, the default values for all parameters were maintained. The word embeddings were initialized with the pre-trained GloVe vectors of 100 dimensions \citep{pennington2014glove} and the character-based embeddings were trained on the corpus at hand (25 dimensions). 

\subsection{Relation Classification}

The second information extraction task we analyze is classifying semantic relations in sentences. As a state-of-the art relation classification method we use the winning system from SemEval 2018 \cite{rotsztejn2018eth}, which combines convolutional and recurrent neural networks to leverage the best architecture for different sentence lengths. We consider the following 11 relation types: \textit{award, employer, education, founder, visited, wife, political-affiliation, nationality, job-title, birth-place} and \textit{death-place}. We use the annotations provided by \citet{culotta2006integrating}.

\paragraph{Features} For this task, we employed the 5 gaze features on word-level provided in the ZuCo data: number of fixations, first fixation duration, total reading time, gaze duration and go-past time. The eye-tracking feature values were normalized over all occurrences in the corpus. The EEG features were extracted by averaging the 105 electrode values over all fixations for each word and then normalized. All word features in a sentence were concatenated and finally padded to the maximum sentence length. The eye-tracking and/or EEG feature vectors were appended to the word embeddings.

\paragraph{Experiments} We performed 5-fold cross validation over 566 samples (sentences can include more than one relation type). We split the data into 80\% training data and 20\% test data.
Due to the small size of the dataset, we used the same preprocessing steps and parameters as proposed by the SemEval 2018 system. The word embeddings were initialized with the pre-trained GloVe vectors of 300 dimensions.

\begin{table*}[t]
\small
\centering
\begin{tabular}{|l|lll|lll|lll|lll|}
\hline
 & \multicolumn{3}{c|}{\textbf{NER}} & \multicolumn{3}{c|}{\textbf{RelClass}} & \multicolumn{3}{c|}{\textbf{Sentiment (2)}} & \multicolumn{3}{c|}{\textbf{Sentiment (3)}} \\\hline \hline
 & P & R & F1 & P & R & F1 & P & R & F1 & P & R & F1 \\\hline 
baseline & 84.5 & 81.7 & 82.9 & 62.6 & 56.6 & 57.7 & 82.5 & 82.5 & 82.5 & 57.1 & 57.6 & 57.2 \\
gaze & 86.2 & 84.3 & \textbf{85.1}** & 65.1 & 61.9 & 62.0** & 84.7 & 84.6 & \textbf{84.6}** & 61.4 & 61.7 & \textbf{61.5}** \\
EEG & 86.7 & 81.5 & 83.9* & 68.3 & 64.8 & \textbf{65.1}** & 83.6 & 83.6 & 83.6** & 60.5 & 60.2 & 60.3** \\
gaze+EEG & 85.1 & 83.2 & 84.0** & 66.3 & 59.3 & 60.8** & 84.3 & 84.3 & 84.3** & 59.8 & 60.0 & 59.8** \\\hline
\end{tabular}
\caption{Precision (P), recall (R) and F1-score (F1) for the four tasks augmented with gaze features, EEG features, and both. Significance is indicated with the asterisks: * = p\textless0.01, ** = p\textless0.0008 (Bonferroni method).}
\label{features-results}
\vspace{-3mm}
\end{table*}

\subsection{Sentiment Analysis}

The third NLP task we choose for this work is sentiment analysis. Based on the analysis by \citet{barnes2017assessing}, we implemented a bidirectional LSTM with an attention layer for the classification of sentence-level sentiment labels.

\paragraph{Features} Analogous to the relation classification, the 5 word-level eye-tracking features were normalized and concatenated before being appended to the sentence embeddings. The raw EEG data (105 electrode values per word) were averaged and normalized. 

\paragraph{Experiments} 10-fold cross validation was performed over the 400 sentences with available sentiment labels from ZuCo (123 neutral, 137 negative and 140 positive sentences). We test ternary classification as well as binary classification. For the latter, we remove all neutral sentences from the training data. Word embeddings were initialized with pre-trained vectors of 300 dimensions \cite{mikolov2013efficient}. All models are trained for 10 epochs with batch sizes of 32. The initial learning rate is set to 0.001. It was halved every 3 passes or every 10 passes, for binary classification and ternary classification respectively (due to the larger training set).


\section{Evaluation}
\vspace{-2mm}
For each information extraction task described in the previous section we trained baseline models, models augmented with gaze features, with EEG features, and with both. All the baseline models were trained solely on textual information (i.e. word embeddings without any gaze or EEG features). We trained single-subject models and models in which the features values are averaged over all subjects.

The results of the averaged models are shown in Table \ref{features-results}. We observe consistent improvements over the baselines for all tasks when augmented with cognitive features. The models with gaze features, EEG features and the combination thereof all outperform the baseline. Notably, while the combination of gaze and EEG features also outperform the baseline, they do not improve over using gaze or EEG individually.

We perform statistical significance testing using permutation (as described in \citet{dror2018hitchhiker}) over all tasks. In addition, we apply the conservative Bonferroni correction for multiple hypotheses, where the global null hypothesis is rejected if $p < \alpha / N$, where $N$ is the number of hypotheses \cite{dror2017replicability}. In our setting, $\alpha=0.01$ and $N=12$, accounting for the combination of the 4 tasks and 3 configurations (EEG, gaze, EEG+gaze). The improvements in 11 configurations out of 12 are also statistically significant under the Bonferroni correction. Despite the limited amount of data, this result suggests that augmenting NLP systems with cognitive features is a generalizable approach.

\begin{figure*}
    \centering
    \includegraphics[scale=0.25]{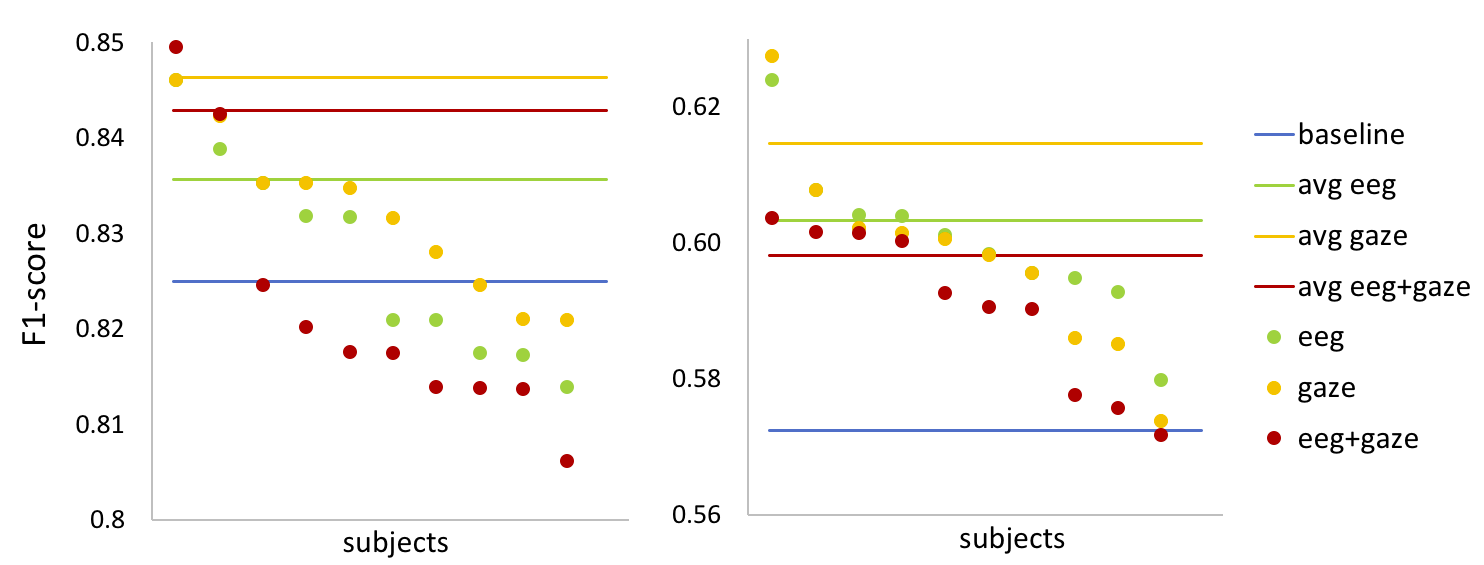}
    \caption{Comparison of single-subject models and features averaged over all subject for both binary sentiment classification (left) and ternary sentiment classification (right). Each dot represents a single subject model, each line an averaged feature model. Note that the best-performing subject for gaze is not necessarily the same subject as for the best EEG model.}
    \label{fig:subjects}
    \vspace{-2mm}
\end{figure*}

\paragraph{Subject analysis} In an additional analysis we also evaluate the single-subject models to test the robustness of averaging the feature values over all readers. By the example of binary and ternary sentiment analysis, Figure \ref{fig:subjects} depicts the variability of the results between the subjects. In contrast to the averaged models, the best subject for binary sentiment classification reaches an F1-score of 85\% with the combination of gaze and EEG data.
Moreover, it shows how the averaged models perform almost as good as the best subject. Note that the best-performing subject for gaze is not necessarily the same subject as for the best EEG model. We also trained models that only take into account the feature values of the five best subjects. However, when averaging over \textit{all} subjects, the signal-to-noise ratio is higher and provides better results than training on the best five subjects only.  While previous research had shown the same effect for using eye-tracking data from multiple subjects in NLP, this had no yet been shown for EEG data. 


\subsection{No real-time recorded data required}\label{sec:type}

While adding these cognitive features to a system show the potential of this type of data, it is not very practical if real-time recordings of EEG and/or eye-tracking are required at prediction time. Following \citet{barrett2016weakly}, we evaluate feature aggregation on word-type level. This means that all cognitive features are averaged over the word occurrences. As a result a lexicon of lower-cased word types with their averaged gaze and EEG feature values was compiled. Words in the training data as well as in the test set are assigned these features if the words occurs in the type-aggregated lexicon or receives unknown features values otherwise. Thus, recorded human data is not required at test time.

We evaluate the concept of type aggregation on the tasks described above. We choose 3 benchmark datasets and add the aggregated EEG and/or eye-tracking features to words occurring in ZuCo. For NER we use the CoNLL-2003 corpus \cite{tjong2003introduction}, for relation classification we use the full Wikipedia dataset provided by \cite{culotta2006integrating} and for sentiment analysis we use the Stanford Sentiment Treebank (SST). The same experiment settings as above were applied here. To avoid overfitting we did not use the official train/test splits but performed cross validation.

Table \ref{type} shows the details about these datasets and the results. We can observe a consistent improvement using type-aggregated gaze features. However, the effect of type-aggregated EEG features is mixed.

\begin{table*}[t]
\small
\centering
\begin{tabular}{|l|cccc|}
\hline
 & \textbf{NER} & \textbf{RelClass} & \textbf{Sentiment (2)} & \textbf{Sentiment (3)} \\\hline\hline
corpus & \textit{CoNLL-2003} & \textit{Wikipedia} & \textit{SST} & \textit{SST}\\
tokens & 302811 & 32953 & 165165 & 202125 \\
sentences & 22137 & 1794 & 9612 & 11853 \\
unknown tokens & 41.09\% & 30.31\% & 26.02\% & 25.96\% \\\hline\hline
baseline & 94.02 & 76.94 & 82.01 & 57.13 \\
gaze & 94.41** & \textbf{77.85}** & 81.64 & \textbf{57.48}** \\
EEG & 94.58** & 76.40 & 80.07 & 54.27 \\
gaze + EEG & \textbf{94.63}** & 77.01 & 79.74 & 54.80 \\\hline
\end{tabular}
\caption{The top part shows the size of the datasets used for the type-aggregation experiments, including the percentage of unknown tokens, i.e. tokens not in the lexicon of aggregated type features. The bottom part shows F1-scores of type aggregation on external benchmark corpora. Significance is indicated with the asterisks: * = p\textless0.01, ** = p\textless0.0008 (Bonferroni method).}
\label{type}
\vspace{-2mm}
\end{table*}



Type aggregation shows not only that recorded gaze or EEG data is not necessary at test time, but also that improvements can be achieved with human data without requiring large quantities of recorded data.



\section{Multi-task learning}\label{sec:mtl}
\vspace{-0.2em}
\begin{table}[h]
\small
\centering
\begin{tabular}{|c|l|c|}
\hline
\textbf{main task} & \textbf{aux task(s)} & \textbf{accuracy} \\\hline\hline
 & - & 87.34 \\
 & freq & 91.29 \\
NER & FFD & 87.34 \\
 & FFD freq & \textbf{91.87} \\
 & EEG$_a$ & 87.31 \\
 & EEG$_a$ freq & 91.79 \\\hline
 & - & 60.99 \\
 & freq & 61.15 \\
Sentiment & TRT & 61.31 \\
binary & TRT freq & 61.13 \\
 & EEG$_b$ & 61.01 \\
 & EEG$_b$ freq & \textbf{61.56} \\\hline
 & - & 61.03 \\
 & freq & 61.02 \\
Sentiment & FFD & 61.05 \\
ternary & FFD freq & 61.10 \\
 & EEG$_t$ & 61.05 \\
 & EEG$_t$ freq & \textbf{61.17} \\\hline
\end{tabular}
\caption{Results of the multi-task learning experiments on NER, binary and ternary sentiment analysis.}
\label{mtl-results}
\vspace{-2mm}
\end{table}

We further investigate multi-task learning (MTL) as an additional machine learning strategy to benefit from cognitive features. The intuition behind MTL is that training signals of one task, the auxiliary task, improves the performance of the main task, by sharing information throughout the training process. In our case, we learn gaze and EEG features as auxiliary tasks to improve the main NLP task. 

In previous work, it has been shown that MTL can be used successfully for sequence labelling tasks \cite{bingel2017identifying} due to some compelling benefits, including its potential to efficiently regularize models and to reduce the need for labeled data. Moreover, gaze duration has been predicted as an auxiliary task to improve sentence compression \cite{klerke2016improving}, and to better predict the readability of texts \cite{gonzalez2018learning}. To the best of our knowledge, EEG features have not been used in MTL to improve NLP tasks.

In multi-task learning it is important that the tasks that are learned simultaneously are related to a certain extent \cite{caruana1997multitask,collobert2011natural}. Assuming that the cognitive processes in the human brain during reading are related, there should be a gain from training on gaze and EEG data when learning to extract information from text. Thus, we assess the hypothesis that MTL might also be useful in our scenario.

\paragraph{Experiments} 

We utilized the Sluice networks \cite{ruder2017learning}, where the network learns to which extent the layers are shared between the tasks. Thus, we re-formulated the sentiment analysis as sequence labelling tasks on phrase level. For binary sentiment analysis, the classes NEUTRAL and NOT-NEUTRAL were predicted. We did not have to modify the named entity recognition task and the relation classification was not tested since only sentence level labels are available.

We ran 5-fold cross validation for all experiments over the same data as described in Section \ref{data}. As our baselines we used single-task learning and learning word frequency as an auxiliary task to an NLP task. Word frequencies were extracted from the British National Corpus \cite{kilgarriff1995bnc}. The experiments ran with the default settings recommended by \cite{ruder2017learning}. In accordance to their results, the Sluice networks yielded consistently higher results than hard parameter sharing.

As a main task the network learned to predict NER, binary or ternary sentiment labels. As auxiliary tasks the network learned a single gaze or EEG feature. We used five eye-tracking features: number of fixations (NFIX), mean fixation duration (MFD), first fixation duration (FFD), total reading time (TRT), and fixation probability (FIXP). Additionally, we tested four EEG features, one for each combined frequency band: EEG$_t$ (i.e. the average values of \textit{theta1} and \textit{theta2}), EEG$_a$, EEG$_b$, EEG$_g$. The features were discretized and binned.

\paragraph{Results}

\begin{table*}[h]
\small
\centering
\begin{tabular}{|l|lllll|llll|}
\hline
 & \multicolumn{5}{c|}{gaze features} & \multicolumn{4}{c|}{EEG features}  \\\hline \hline
 & \textbf{NFIX} & \textbf{MFD} & \textbf{FFD} & \textbf{TRT} & \textbf{FIXP} & \textbf{EEG$_t$} & \textbf{EEG$_a$} & \textbf{EEG$_b$} &\textbf{EEG$_g$} \\\hline
- & 64.14 & 84.60 & 55.21 & 65.04 & 46.66 & 40.67 & 36.14 & 39.50 & 30.48 \\
freq & 71.01 & 84.64 & \textbf{63.68} & 71.99 & \textbf{56.34} & 53.36 & \textbf{49.75} & \textbf{52.79} & \textbf{41.34}\\
gaze & \textbf{71.34} & \textbf{84.78} & 63.60 & \textbf{72.20} & 55.77 & \textbf{53.53} & 49.38 & 52.58 & 40.95\\
EEG & 71.15 & 84.64 & 62.10 & 72.03 & 55.63 & 53.47 & 46.77 & 52.54 & 37.27 \\\hline
\end{tabular}
\caption{Learning cognitive features in an MTL setting. Columns = main tasks, rows = auxiliary tasks.}
\label{mtl}
\end{table*}

Table \ref{mtl-results} shows the results of these experiments. Note that only the best feature combinations are included in the table. Learning word frequency as an auxiliary task is a strong baseline. Learning gaze and EEG features as auxiliary tasks does not improve the performance over the single-task baseline for NER and only minimally for sentiment analysis. Learning two auxiliary tasks, a gaze of EEG feature \textit{and} word frequency in parallel yields modest improvements over the frequency baseline. 

Adding further auxiliary tasks with additional gaze or EEG features did not yield better results. Moreover, the combination of learning gaze and brain activity features did also not bring further improvements.

As we know that gaze and frequency band EEG features represent different cognitive processes involved in reading, our main and auxiliary tasks should in fact be related. However, it seems like the noise-to-signal ratio in the EEG features is too high to achieve significant results. As stated by \citet{gonzalez2018learning}, it is important to establish whether the same feature representation can yield good results for all tasks independently. To gain further insights into these results, we analyze how well these human features can be learned.

\subsection{Learning cognitive features}

Using the same experiment setting as for the above described MTL experiments, we first trained single-task baselines for each of the gaze and EEG features. Then, we trained each gaze feature in 3 MTL settings: (1) word frequency as an auxiliary task, (2) the remaining gaze features as parallel auxiliary tasks and (3) the EEG features as parallel auxiliary tasks. The same applies to EEG features as main tasks. The results in Table \ref{mtl} show that gaze features have far higher baselines than EEG features. Presumably EEG is harder to learn because it has larger variance in the data. Moreover, while the eye-tracking data is limited to the visual component of the cognitive processes, EEG data additionally contains a motor component and a semantic component during the reading process. 

Learning word frequency as an auxiliary task considerably helps all gaze and EEG features. The known correlation between eye-tracking and word frequency \cite{rayner1986lexical} is clearly beneficial for learning gaze features. Moreover, a frequency effect can also be found in \textit{early} EEG signals, i.e. during the first 200ms of reading a word \cite{hauk2004effects}.

\section{Discussion}
\vspace{-0.2em}
In accordance with previous work (e.g. \citet{barrett2016weakly,mishra2018cognitively}), we showed consistent improvements when using gaze data in a range of information extraction tasks, with recorded token-level features and with type-aggregated features on benchmark corpora. The patterns in the results are less consistent when enhancing NLP methods with EEG signals. While we can still show significant improvements over the baseline models, in general the models leveraging EEG features yield lower performance than the ones with gaze features. A plausible explanation for this is that the combination of gaze and EEG features decreases the signal-to-noise ratio even more than for only one type of cognitive data. Another interpretation is that the eye-tracking and EEG signal contain information that is (too) similar. Thus, the combination does not improve yield better results.

Consequently, there are some open questions: How can EEG signals be preprocessed and de-noised more efficiently for NLP tasks?
How can EEG and eye-tracking (and other cognitive processing signals or fortuitous data \cite{plank2016non}) be combined more effectively to improve NLP applications?

The models leveraging type-aggregated cognitive features show that improvements can be achieved without requiring large amounts of recorded data and provide evidence that this type of data can be generalized on word type level. Although these results indicate that huge amounts of recorded data are not necessary for performance gains, one of the limitations of this work is the effort of collecting cognitive processing signals from humans. However, webcam-based eye-trackers (e.g. \citet{papoutsaki2016webgazer}) and commercially available EEG devices (e.g \citet{stytsenko2011evaluation}) are becoming more accurate and user-friendly.

Finally, the multi-task learning experiments provide insights into the correlation of learning NLP tasks together with word frequency and cognitive features. While the results are not as promising as our main experiments, it reveals qualities of the individual gaze and EEG features.
For future work, a possible approach to combine the potential of exceptionally good single-subject models and multi-task learning, would be to learn gaze and/or EEG features from multiple subjects at the same time. This has been shown to improve accuracy on brain-computer interface tasks and helps to further reduce the variability between subjects \cite{panagopoulos2017multi}.

One of the challenges of NLP is to learn as much as possible from limited resources. Using cognitive language processing data may allow us take a step towards meta-reasoning, the process of discovering the cognitive processes that are used to tackle a task in the human brain \cite{griffiths2019doing}, and in turn be able to improve NLP.

\section{Conclusion}
\vspace{-0.2em}
We presented an extensive study of improving NLP tasks with eye-tracking and electroencephalography data as instances of cognitive processing signals. We showed how adding gaze and/or EEG features to a range of information extraction tasks, namely named entity recognition, relation classification and sentiment analysis, yields significant improvements over the baselines. Moreover, we showed how these features can be generalized at word type-level so that no recorded data is required during prediction time. Finally, we explored a multi-task learning setting to simultaneously learn NLP tasks and cognitive features.

In conclusion, the gaze and EEG signals of humans reading text, even though noisy and available in limited amounts, show great potential in improving NLP tasks and facilitate insights into language processing which can be applied to NLP, but need to be investigated in more depth.

\bibliography{acl2019}
\bibliographystyle{acl_natbib}

\end{document}